\documentclass[runningheads]{llncs}

\usepackage{eccv}

\usepackage{eccvabbrv}

\usepackage{graphicx}
\usepackage{booktabs}
\usepackage[dvipsnames]{xcolor}
\usepackage{colortbl}
\usepackage{bm}
\usepackage{multirow}

\usepackage[accsupp]{axessibility}  

\usepackage[pagebackref,breaklinks,colorlinks,citecolor=eccvblue]{hyperref}

\usepackage{orcidlink}

\begin{document}

\title{Idempotent Unsupervised Representation Learning for Skeleton-Based Action Recognition} 

\titlerunning{Idempotent Unsupervised Representation}

\author{Lilang Lin\orcidlink{0000-0002-3229-4096} \and
Lehong Wu\orcidlink{0009-0008-1155-5577} \and
Jiahang Zhang\orcidlink{0000-0001-5039-8916} \and
Jiaying Liu\thanks{Corresponding author.}\orcidlink{0000-0002-0468-9576}}

\authorrunning{Lin et al.}

\institute{Wangxuan Institute of Computer Technology, Peking University\\
\email{\{linlilang, zjh2020, liujiaying\}@pku.edu.cn\\
aladonwlh@pku.stu.edu.cn}}

\maketitle

\begin{abstract}
Generative models, as a powerful technique for generation, also gradually become a critical tool for recognition tasks.
However, in skeleton-based action recognition, the features obtained from existing pre-trained generative methods contain redundant information unrelated to recognition, which contradicts the nature of the skeleton's spatially sparse and temporally consistent properties, leading to undesirable performance.
To address this challenge, we make efforts to bridge the gap in theory and methodology and propose a novel skeleton-based idempotent generative model (IGM) for unsupervised representation learning.
More specifically, we first theoretically demonstrate the equivalence between generative models and maximum entropy coding, which demonstrates a potential route that makes the features of generative models more compact by introducing contrastive learning.
To this end, we introduce the idempotency constraint to form a stronger consistency regularization in the feature space, to push the features only to maintain the critical information of motion semantics for the recognition task.
%
%
Our extensive experiments on benchmark datasets, NTU RGB+D and PKUMMD, demonstrate the effectiveness of our proposed method. On the NTU 60 xsub dataset, we observe a performance improvement from 84.6$\%$ to 86.2$\%$. Furthermore, in zero-shot adaptation scenarios, our model demonstrates significant efficacy by achieving promising results in cases that were previously unrecognizable. Our project is available at \url{https://github.com/LanglandsLin/IGM}.
\end{abstract}

\section{Introduction}
\label{sec:intro}

Skeletons represent human joints through 3D coordinate locations, providing a compact and efficient modality of representing human motion compared to RGB videos and depth data.
Owing to their simplicity and superior discriminative capabilities for analysis, skeleton representations have been extensively employed in the field of action recognition tasks~\cite{zhang2020context,liu2020disentangling,zhang2020semantics,song2020stronger,peng2020learning,su2020predict}.

Supervised skeleton-based action recognition methods~\cite{si2019attention,shi2019two,chen2021channel} have demonstrated remarkable performance. However, they {heavily rely on vast amounts of labeled training data}, the collection of which can be a costly and time-consuming process.
In order to reduce the reliance on fully supervised paradigms, self-supervised learning approaches have been explored in the context of skeleton-based action recognition~\cite{zheng2018unsupervised,lin2020ms2l,su2020predict,thoker2021skeleton}.
%

In the context of self-supervised pretraining paradigms, most methods can be broadly classified into two categories: generative learning-based~\cite{su2020predict,yang2021skeleton,kim2022global} and contrastive learning-based approaches. Generative learning-based methods typically model the spatial-temporal correlations by predicting or reconstructing the masked skeleton data.
With long-term global motion dynamics, Zheng \textit{et al.}~\cite{zheng2018unsupervised} were the pioneers in introducing the concept of reconstructing masked skeleton data.
The structure of Masked Auto-Encoder (MAE) was used by Mao \textit{et al.}~\cite{mao2023masked} to predict the velocity of the masked part thus obtaining motion information modelling.
However, skeleton data is by nature spatially sparse and temporal consistent while MAE's feature preserves too much appearance information, which will interfere with the recognition tasks.

On the other route, contrastive learning-based methods also have recently demonstrated remarkable potential. These methods utilize skeleton transformations to generate positive pairs and aim to maintain consistency in the embedding space.
Rao \textit{et al.}~\cite{rao2021augmented} introduced shearing and cropping as data augmentation techniques.
Guo \textit{et al.}~\cite{guo2021contrastive} extended these efforts by suggesting additional augmentations, such as rotation, masking, and flipping, to further enhance the consistency of contrastive learning. Contrastive learning, aimed at high-level tasks like recognition, often requires data transformation to filter out task-irrelevant information. This process results in a significant loss of information in the extracted features and hampers the ability to capture fine-grained motion details.

However, previous research has typically focused on these two paradigms separately. Their ideas and technical advantages are complementary and can be augmented, which is still under-explored. To address this gap, we first seek theoretical inspiration about the relationship between generative models and contrastive learning. In detail, we find that generative methods are equivalent to maximum entropy coding. This fact naturally inspires building the generative models with the related idempotent constraint to form a novel idempotent generative model, which is exactly equivalent to spectral contrastive learning but with improved recognition capacities.

Building upon this theoretical foundation, we propose a novel idempotent generative model to promote consistency in the feature space. By enforcing idempotence at the feature and distribution levels, our model enriches features with semantic motion information, thereby reducing the domain gap and adapting the generative model better for recognition. 
Moreover, the features of generative models are span on principal components, which easily leads to dimensional collapse, as recognition tasks primarily rely on descriminative local details.
To address this imbalance, we introduce an adapter that fuses encoder and generator features. This integration expands the effective feature dimension of the feature space, facilitating more robust and comprehensive representation.
Our model attains outstanding results through self-supervised learning in comparison to contemporary state-of-the-art methods.

In summary, our contributions {are three-folded}:
\begin{itemize}
\setlength\itemsep{1em}
\item We propose an idempotent generative model to combine the benefits of generative pre-training and contrastive learning, which is inspired by the theoretical fact of their intrinsic correlation. This cooperation makes the model focus on extracting more compact information related to motion semantics, and obtain more powerful high-level representation within the generative model framework.

\item We further propose to utilize a multiple idempotency feature constraint. Through feature and distribution idempotency constraints, the feature consistency is improved, leading to not only improved recognition capture but also the perceptual reconstruction quality of the generative model. 

\item We employ an adapter to fuse the features from the high-level semantic encoder and low-level skeleton generator from different subspaces to expand the representation dimension. Experiments show that our module improves the effective dimension of the feature space and encodes rich information.

\end{itemize}

\section{Related Work}
\subsection{Skeleton-Based Action Recognition}

Skeleton sequences encode the motion trajectories of human joints, representing rich information about human actions. Thus, skeleton data serves as a suitable modality for human action recognition~\cite{zhang2020context,liu2020disentangling,zhang2020semantics,song2020stronger}. Skeleton can be obtained by applying pose estimation algorithms on RGB videos or depth maps~\cite{shotton2013real}. 


Early studies focused on extracting hand-designed spatial and temporal domain features from skeleton sequences for human movement recognition~\cite{vemulapalli2014human,vemulapalli2016rolling,wang2012mining,goutsu2015motion, lv2006recognition}. In later work, efforts were made to model the positional information and higher-order temporal difference information of the human skeleton~\cite{tao2015moving,vemulapalli2014human}. Additionally, graphical models were built by tracking the trajectory of human joints to capture joint information in video sequences~\cite{wang2016graph}. 

Recently, there has been a surge of interest in using graph structures for learning models~\cite{wang2016graph}. Graph Neural Network (GNN) is one such model capturing intra-graph dependencies through information transfer between nodes. Various approaches have been proposed, such as spatio-domain inference networks, recurrent neural networks (RNNs), and graph convolutional networks (GCNs), to exploit graph structures for human action recognition~\cite{si2018skeleton,shi2019skeleton,yan2018spatial}. These models automatically learn spatio-temporal patterns from skeleton data, facilitating strong action generalization. Moreover, attention mechanisms, multiscale aggregation schemes, and lightweight convolution operations have been integrated into GCN-based models to enhance their effectiveness and reduce computational costs~\cite{liu2020disentangling,zhang2020semantics,cheng2020skeleton,song2022constructing}.

\subsection{Self-Supervised Learning}
The self-supervised task aims to extract data features from a large amount of unlabelled data~\cite{erhan2010does}. It can be widely used in semantic segmentation, image classification, action recognition and many other tasks~\cite{jang2018grasp2vec,owens2018audio}. These tasks are mainly classified into methods based on reconstruction and on contrastive learning.

Reconstruction based approach after masking part of the original data, the network is used to reconstruct the masked part of the data. He \etal~\cite{he2022masked} proposed Mased Auto-Encoder~(MAE) to encode the visible patches and decode the visible and masked patches. This approach has been extended to the video domain and has been used in several studies. These methods typically use a visual Transformer as the backbone network in order to perform the mask reconstruction task. Feichtenhofer \etal~\cite{feichtenhofer2022masked} extended the image-based masked auto-encoder to use spatio-temporal learning to randomly mask spatio-temporal segments of a video and learn am auto-encoder for reconstruction at the pixel-level reconstruction. Similarly, in MaskFeat, Wei \etal~\cite{wei2022masked} used several video cubes and utilized the model to predict them using the remaining information.

Contrastive learning pushes pairs of positive sample together while pushing pairs of negative sample further apart. To generate negative samples, contrastive learning pairs anchor frames with frames from other videos. There are various ways of generating positive and negative samples, which is the main factor that distinguishes different contrastive methods.

Most of these methods generate positive and negative samples by different ways in order to minimize and maximise the distance between them respectively. In the image domain, positive samples are usually generated by enhancing the image in different ways~\cite{tian2019contrastive,wu2018unsupervised,bachman2019learning,ye2019unsupervised,isola2015learning}. These enhancements include rotation, cropping, random greyscale and colour change~\cite{chen2020simple}. Scaling these methods in video can be difficult because each video comparison increases the memory required, especially if multiple enhancements are used for multiple positive samples. Another challenge is incorporating the temporal domain into the enhancement. Some methods simply apply the same enhancement in the image to each frame~\cite{tian2020contrastive}. Some methods include additional frame alignments that may be based on the temporal domain~\cite{lorre2020temporal}. Finally, some methods rely on motion and optical flow maps as positive samples~\cite{rai2021cocon}.

\section{Idempotency Generation Network (IGN)}

\subsection{Self-Conditional Generative Models as Maximum Entropy Coding}
Self-conditional generative modeling~\cite{he2022masked} is frequently employed as a pre-training task in self-supervised learning. It is generally structured as an auto-encoder. Formally, given the input skeleton data $\mathbf{x}$, the reconstruction loss is:
\begin{equation}
\mathcal{L} = \mathbb{E}_{\mathbf{x} \sim p_{\mathbf{x}}} \left[\mathcal{D}(g(\mathbf{z}), \mathbf{x})\right] = \mathbb{E}_{\mathbf{x} \sim p_{\mathbf{x}}} \left[\mathbb{E}_{\mathbf{z} \sim p_{\mathbf{z}|\mathbf{x}}}[-\log p(\mathbf{x}|\mathbf{z})]\right] = H(\mathbf{x}|\mathbf{z}),
\end{equation}
where $\mathbf{z} = f(\mathcal{T}(\mathbf{x}))$, $f(\cdot)$ is the encoder, and $\mathcal{T}(\cdot)$ is data transformation. $g(\cdot)$ is the generator. $\mathcal{D}(\cdot,\cdot)$ is the distance. $p_{\mathbf{x}}$ is the data distribution and $ p_{\mathbf{z}|\mathbf{x}}$ is the feature distribution given $\mathbf{x}$. $H(\cdot|\cdot)$ is the conditional entropy. In the context of MAE, this data transformation represents masked data during training. Conversely, in denoising auto-encoders, this transformation signifies adding noise to the input data.

In the context of mutual information, this loss function is equivalent to optimizing the mutual information $I(\mathbf{z}; \mathbf{x})$ between the extracted features $\mathbf{z}$ and the input data $\mathbf{x}$. Based on the relationship between mutual information and entropy, we get
\begin{equation}
I(\mathbf{z}; \mathbf{x}) = H(\mathbf{x}) - H(\mathbf{x}|\mathbf{z}) = H(\mathbf{z}) - H(\mathbf{z}|\mathbf{x}).
\label{eq:entropy}
\end{equation}
Since the entropy of $\mathbf{x}$ remains constant, decreasing the reconstruction loss is akin to increasing the mutual information. Conversely, as the features $\mathbf{z}$ are deterministically derived from the data $\mathbf{x}$ by an encoder $f(\cdot)$, the entropy $H(\mathbf{z}|\mathbf{x})$ tends towards zero. Hence, maximizing the mutual information $I(\mathbf{z}; \mathbf{x})$ is equivalent to maximizing the entropy of the feature space $H(\mathbf{z})$.

Estimating the true distributions $p(\mathbf{z})$ of the representation space is exceedingly challenging. 
Following works~\cite{yu2020learning,liu2022self}, we leverage lossy data coding, a computationally feasible alternative, as a surrogate for the entropy of continuous random variables $H(\mathbf{z})$. This approach involves determining the minimal number of bits required to encode a set of samples \( \mathbf{Z} = [\mathbf{z}^1, \dots, \mathbf{z}^m] \in \mathbb{R}^{d \times m} \) subject to a distortion \( \varepsilon \), as defined by the coding length function below [47, 72]:
\begin{equation}
L = \left(\frac{m + d}{2}\right) \log \det \left(\mathbf{I} + \frac{d}{m\varepsilon^2} \mathbf{Z}^T \mathbf{Z}\right),
\end{equation}
where $\varepsilon$ is the upper bound of the expected decoding error between $\mathbf{z} \in \mathbf{Z}$ and the decoded $\hat{\mathbf{z}}$. $\det(\cdot)$ is the determinant of a matrix. $d$ is the dimension of the feature space. Utilizing the identity \( \det(\exp(\mathbf{A})) = \exp(\text{Tr}(\mathbf{A})) \), we derive \( L = \text{Tr} \left( \mu \log \left( \mathbf{I} + \lambda \mathbf{Z}^T \mathbf{Z} \right) \right) \), where \( \text{Tr} \) denotes the trace of the matrix and $\mu = \frac{m + d}{2}$, $\lambda = \frac{d}{m\varepsilon^2}$. Finally, we apply a Taylor series expansion to the logarithm of the matrix to obtain:
\begin{equation}
L = \text{Tr} \left( \mu \sum_{n=1}^\infty \frac{(-1)^{n-1}}{n} \left(\lambda \mathbf{Z}^T \mathbf{Z} \right)^n \right),
\end{equation}
because the features $\mathbf{z}$ are projected into spherical space $\mathbb{S}^{d-1}$, $\text{Tr}\left( \mu \lambda \mathbf{Z}^T \mathbf{Z} \right) = m \mu \lambda$. Hence, the first term does not contribute to the reconstruction learning. In essence, self-conditional generation primarily diminishes the inter-data similarity within the feature space:
\begin{equation}
L = - \frac{\mu\lambda^2}{2} \text{Tr} \left( \left( \mathbf{Z}^T \mathbf{Z} \right)^2 \right) - \mathbf{R} = - \frac{\mu\lambda^2}{2} \sum_{i=1}^m \sum_{j=1}^m (\mathbf{z}_i^T \mathbf{z}_j)^2 - \mathbf{R},
\end{equation}
where $\mathbf{R} = \sum_{n=3}^\infty \frac{(-1)^{n}\mu \lambda^n}{n} \text{Tr} \left(\left(\mathbf{Z}^T \mathbf{Z} \right)^n \right)$.

\subsection{Idempotent Generative Models as Spectral Contrastive Learning}
The idempotence of a self-conditional generative model refers to its stability in re-encoding~\cite{xu2024idempotence}. More precisely, if we denote the original data as \( \mathbf{x} \), the encoder as \( f(\cdot) \), the encoded feature as \( \mathbf{z} = f(\mathbf{x}) \), the decoder as \( g(\cdot) \), and the reconstruction as \( \hat{\mathbf{x}} = g(\mathbf{z}) \), then the self-conditional generative model is considered idempotent:
\begin{equation}
f(\hat{\mathbf{x}}) = \mathbf{z} \quad \text{or} \quad g(f(\hat{\mathbf{x}})) = \hat{\mathbf{x}}.
\end{equation}
Idempotence is frequently employed in the generative domain to augment the perceptual loss of generated images. The idempotent loss is formulated as:
\begin{equation}
\mathcal{L}_{\text{ide}} = \|f(\hat{\mathbf{x}}) - \mathbf{z}\|^2 = 2 - 2 f(\hat{\mathbf{x}})^T f(\mathbf{x}),
\end{equation}
where $\mathbf{z}^T \mathbf{z} = 1$ because we normalize the feature space. Therefore, the idempotent generative model maximizes the entropy of the feature space while simultaneously minimizing the feature distance between the data and the generated data. The total loss of the idempotent generative model is expressed as:
\begin{equation}
\begin{aligned}
\mathcal{L} &= \mathcal{L}_{\text{ide}} - L = -2 \sum_{\mathbf{x}, \hat{\mathbf{x}}} p(\mathbf{x}, \hat{\mathbf{x}}) f(\hat{\mathbf{x}}_i)^T f(\mathbf{x}_i) + \sum_{\mathbf{x}, \mathbf{x}'} p(\mathbf{x})p(\mathbf{x}') \left(f(\mathbf{x})^T f(\mathbf{x}')\right)^2 
+ \mathbf{R}\\
& = - 2 \mathbb{E}_{(\mathbf{x}, \hat{\mathbf{x}}) \sim p(\mathbf{x}, \hat{\mathbf{x}})}\left[f(\hat{\mathbf{x}})^T f(\mathbf{x})\right] + \mathbb{E}_{(\mathbf{x}, \mathbf{x}') \sim p(\mathbf{x})p(\mathbf{x}')}\left[\left(f(\mathbf{x})^T f(\mathbf{x}')\right)^2\right] + \mathbf{R}\\
&= -2 \text{Tr}\left( \mathbf{F} \mathbf{A} \mathbf{F}^T \right) + \text{Tr} \left( \left( \mathbf{F}^T \mathbf{F} \right)^2 \right) 
+ \mathbf{R} = 2 \text{Tr}\left( \mathbf{F} \mathbf{L} \mathbf{F}^T \right) + \text{Tr} \left( \left( \mathbf{F}^T \mathbf{F} \right)^2 \right) + \mathbf{R} + \mathbf{C}\\
&= \|\mathbf{A} - \mathbf{F}^T \mathbf{F}\|_F^2 + \mathbf{R} + \mathbf{C},
\end{aligned}
\end{equation}
where $\mathbf{A} \in \mathbb{R}^{m \times m}$ is the adjacency matrix defined by the data generation and $\mathbf{C}$ is a constant. $\mathbf{F} = \mathbf{Z} \text{diag}(\sqrt{p(\mathbf{x})})$ The weights $\mathbf{A}_{\mathbf{x},\hat{\mathbf{x}}} = \frac{p(\mathbf{x},\hat{\mathbf{x}})}{\sqrt{p(\mathbf{x})p(\hat{\mathbf{x}})}}$. $\mathbf{L} = \mathbf{I} - \mathbf{A}$ is the Laplacian matrix.
This demonstrates its equivalence to spectral contrastive learning. And the advantage of our approach over spectral contrastive learning is that we additionally optimise the residual term $\mathbf{R}$ to capture higher order information.

Further, we exploit data idempotence and feature idempotence to enhance representation learning and action generation. The unified generative-perceptual model contains both an encoder $f(\cdot)$ and generator $g(\cdot)$. And our idempotency constraints pay attention to both the data and the feature distributions, which improves both generation and feature learning.
\begin{equation}
(g \circ f)(\mathbf{x}) = \mathbf{x} \quad \text{and} \quad (f \circ g)(\mathbf{z}) = \mathbf{z}.
\end{equation}

\subsection{Relationship to Masked Auto-Encoder}
As mentioned in Eq. \ref{eq:entropy}, the generative network without idempotent constraints has a conditional entropy $H(\mathbf{z}|\mathbf{x})$ of 0 because the encoding process is deterministic. Idempotent generative networks, on the other hand, treat features $\mathbf{z}$ as a random variable sampled from the distribution of features across all of the generated data $\hat{\mathbf{x}}$ for the same data $\mathbf{x}$, thus transforming into a non-deterministic process:
\begin{equation}
\mathbf{z} = f(\hat{\mathbf{x}}), \quad \hat{\mathbf{x}} \sim G(\mathbf{x}),
\end{equation}
where $G(\cdot)$ is the generation process. Therefore, idempotent constraints are essentially about diminishing conditional entropy $H(\mathbf{z}|\mathbf{x})$, which in turn maximizes the mutual information between features and data. 

In contrast, methods like MAE implicitly prioritize maximizing feature similarity across masked samples of the same data:
\begin{equation}
\mathbf{z} = f(\hat{\mathbf{x}}), \quad \hat{\mathbf{x}} \sim M(\mathbf{x}),
\end{equation}
where $M(\cdot)$ is the random masking process. 
Consequently, features from two distinct data that undergo similar transformed or generated data are clustered into the same class. However, the data obtained through data transformation may not be the real data and thus far from the real data distribution. 

\subsection{Relationship to Downstream Tasks} 
Through the analysis of previous work~\cite{haochen2106provable,zhang2022mask,wang2023message,guo2023contranorm,du2023role} on spectral contrastive learning, the error rate $P_e = P[\phi(\mathbf{x}) \not = y_{\mathbf{x}}]$ of the downstream linear evaluation $\phi(\cdot)$ can be bounded by the generated adjacency matrix $\mathbf{A}$ and clustering error probabilities $\alpha = P[y_{\mathbf{x}} \not = y_{\hat{\mathbf{x}}}]$:
\begin{theorem}
If $\lambda_1 \geq \lambda_2 \geq \dots \geq \lambda_m$ are the eigenvalues of $\mathbf{A}$, and if the clustering purity is $1 - \alpha$, we obtain:
\begin{equation}
P_e \leq c_1 \sum_{i=d+1}^m \lambda_i^2 + c_2 \alpha,
\end{equation}
where $c_1, c_2$ are some constants.
\end{theorem} 
This theorem illustrates the constraints on accuracy imposed by the purity $1 - \alpha$. A large purity and a small  number of clusters result in a low error rate. When the diversity in the generated data is insufficient, the sum of small singular values of the adjacency matrix become large, resulting in less tightly clustered groups. Conversely, excessive diversity in the generated data may compromise the preservation of motion information, thereby increasing the error rate in clustering and undermining overall clustering effectiveness.

Therefore, to make the feature space of the idempotent generative model more capable of clustering, it is necessary to increase the diversity of the generated data for a stronger feature consistency constraint. However, a paradox is demonstrated here. Ordinary generative processes result in limited diversity under self-conditional generation due to constraints on the distance between the generated data and the original data. So in order to simultaneously obtain diverse and motion semantics preserving generated data, we propose an idempotent self-conditional generation model based on the diffusion generation model. The diversity of the generated data is provided by the noise sampling process of the diffusion model.

\begin{figure*}[tb]
\begin{center}
\includegraphics[width=\textwidth]{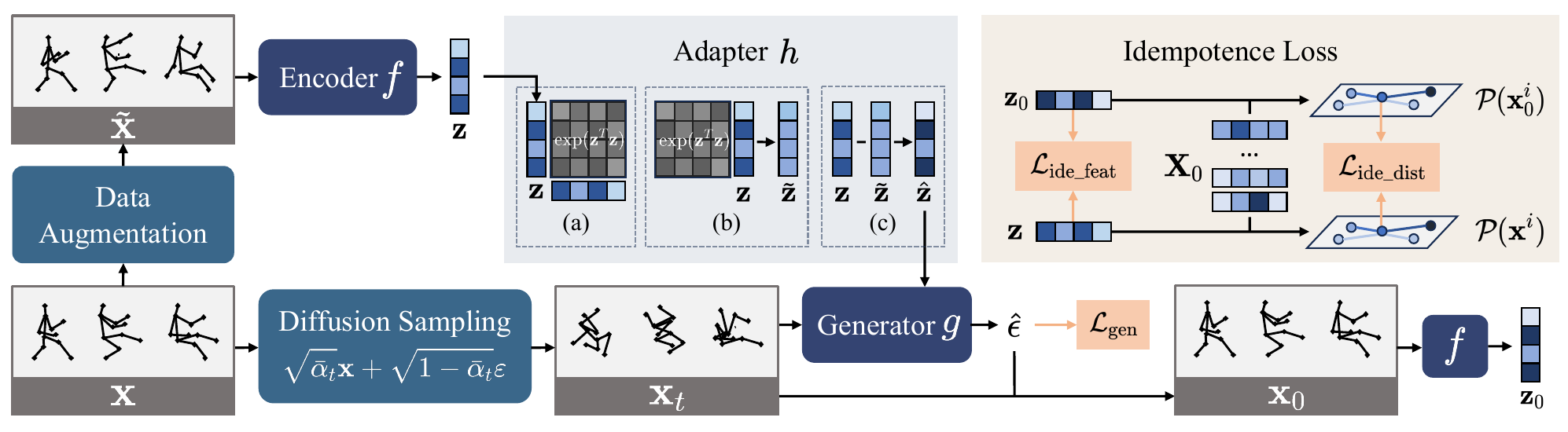}
\end{center}
\caption{
We perform data augmentations on the data first and then obtain the conditional features through the encoder $f(\cdot)$. The noise skeleton is then obtained using Diffusion Sampling. The noise skeleton and conditions are fed into the generator $g(\cdot)$ for denoising. 
The adapter $h(\cdot)$ plays a pivotal role in projecting and fusing the features extracted by the encoder $f(\cdot)$ into the generator's feature space for use as conditions. In the adapter, (a) involves computing similarity using spatio-temporal tokens within the sequence. (b) calculates similar tokens based on the similarity of each token. (c) entails de-correlation by subtracting similar tokens. This integration expands the effective feature dimension of the feature space, facilitating more robust and comprehensive representation. We utilize two losses in our model: Diffusion's noise prediction loss and idempotent feature constraints, which respectively constrain feature similarity and distributional similarity. Thus, the feature consistency is improved, leading to not only improved recognition capture but also the perceptual reconstruction quality of the generative model. 
}
\label{fig:overview}
\end{figure*}

\subsection{Idempotent Diffusion Generation Model}
Our model consists of three parts, an encoder $f(\cdot)$, a generator $g(\cdot)$ and an adapter $h(\cdot)$. The encoder $f(\cdot)$ extracts features $\mathbf{z}$ as conditions for the generator $g(\cdot)$ and also as inputs to the downstream task classifier $\phi(\cdot)$. And the generator $g(\cdot)$ reconstructs the skeleton data based on the features. The adapter $h(\cdot)$, in turn, is responsible for projecting and fusing the features extracted by the encoder $f(\cdot)$ into the generator's feature space to be used as conditions. 


\noindent \textbf{Encoder $f(\cdot)$:} We start by applying some data augmentations to the data $\mathbf{x}$ to obtain data $\Tilde{\mathbf{x}}$ for increasing diversity. Then, spatio-temporal position embeddings $P_t$ and $P_v$ are added after projection to the feature space by linear projection:
\begin{equation}
\mathbf{z} = \text{LinearProj}(\Tilde{\mathbf{x}}) + P_t + P_v.
\end{equation}
Following that, layers of vanilla transformer blocks are employed to extract latent representations $\mathbf{z}$. Each block consists of a multi-head self-attention (MSA) module and a feed-forward network (FFN) module. Residual connections are utilized within each module, which are then followed by layer normalization (LN).


\noindent \textbf{Generator $g(\cdot)$:} The generator $g(\cdot)$ and encoder $f(\cdot)$ maintain the same structure. But the input is the noise data $\mathbf{x}_t$ obtained by sampling in the diffusion. The generator $g(\cdot)$ predicts the noise magnitude by taking noise data $\mathbf{x}_t$ and feature conditions $\mathbf{z}$ as inputs.


\noindent \textbf{Adapter $h(\cdot)$:} The adapter $h(\cdot)$ merges the features extracted by the encoder $f(\cdot)$ into the generator $g(\cdot)$. This is necessary because high-level tasks like recognition operate in a different feature space compared to low-level tasks like generation. 
Recognition tasks necessitate capturing high-frequency action movements while generation primarily focuses on optimizing principal component space (with large singular values), such as walking or waving, which rely more on bottom component subspace like velocity. Thus, we introduce a feature fusion method that decouples principal and bottom component subspace, allowing the encoder features to focus more on high-frequency information, making them more suitable for high-level tasks such as action recognition. These features are then injected into the bottom component feature space of the generator.


\noindent$\bullet$ \textbf{Manifold Decoupled Feature Fusion Module}: To derive discriminative features for use as semantic guides, we draw inspiration from negative samples in contrastive learning. We assume that regions with motion semantics have the lowest similarity to other regions in the same sequence. $\mathbf{z} = [\mathbf{z}_1,\dots,\mathbf{z}_l] \in \mathbb{R}^{d \times l}$, where $l$ is the length of tokens of $\mathbf{z}$. The uniformity loss in contrastive learning is:
\begin{equation}
\mathcal{L}_{\text{uni}} = \mathbb{E}_{\mathbf{z}_i} \left[\log \mathbb{E}_{\mathbf{z}_j} \left[\exp\left({\mathbf{z}_i}^T \mathbf{z}_j\right)\right]\right] = \text{Tr}\left(\log\left(\deg\left(\exp(\mathbf{z}^T \mathbf{z})\right)\right)\right),
\end{equation}
the derivative of $\mathcal{L}_{\text{uni}}$ is as follows:
\begin{equation}
\hat{\mathbf{z}} \Leftarrow \mathbf{z} - \eta \frac{\partial \mathcal{L}_{\text{uni}}}{\partial \mathbf{z}} = \mathbf{z} - \eta {\mathbf{D}'}^{-1} \mathbf{A}' \mathbf{z},
\end{equation}
where $\mathbf{A}' = \exp({\mathbf{z}}^T \mathbf{z})$ and $\mathbf{D}' = \text{deg}(\mathbf{A}')$. ${\mathbf{D}'}^{-1} \mathbf{A}' = \text{SoftMax}({\mathbf{z}}^T \mathbf{z})$. $\hat{\mathbf{z}}$ removes low-frequency information. Based on this analysis, we extract the high-frequency information of the features as semantic information:
\begin{equation}
\hat{\mathbf{z}} \Leftarrow (1 + \eta)\mathbf{z} - \eta \text{SoftMax}({\mathbf{z}}^T \mathbf{z})\mathbf{z}.
\end{equation}
Through this high-pass filtering, we filter out some low-frequency information of principal component space such as the mean value in the sequences, which is not very meaningful for recognition, and retain the semantic information, which is more important for recognition. This module also mitigates dimensionality collapse, making features more informative.

We then fuse the features into the generator by replacing LayerNorm (LN) with Adaptive LayerNorm (AdaLN) with the following equation:
\begin{equation}
\text{AdaLN}(\mathbf{h},\hat{\mathbf{z}},t) = \hat{\mathbf{z}}_s \cdot \left(\mathbf{t}_s \cdot \text{LN}(\mathbf{h}) + \mathbf{t}_b\right) + \hat{\mathbf{z}}_b 
\end{equation}
where $\mathbf{h}$ represents the hidden representation of the generator, $(\mathbf{t}_s,\mathbf{t}_b)$ and $(\hat{\mathbf{z}}_s,\hat{\mathbf{z}}_b)$ are obtained from linear projection of timestep embedding $t$ and high-frequency condition $\hat{\mathbf{z}}$, respectively. Through AdaLN layers, the condition $\hat{\mathbf{z}}$ guides the denoising process by scaling and shifting the normalized hidden representation.


\noindent \textbf{Idempotence Generation Loss:}
Our loss function comprises two components: the noise prediction loss of the diffusion model and the idempotency constraint.

\noindent$\bullet$ \textbf{Noise Prediction Loss}: The diffusion model is trained by predicting the noise from the input noise data:
\begin{equation}
\begin{aligned}
\mathcal{L}_{\text{gen}} &= \|g(\mathbf{x}_t, h(\mathbf{z}), t) - \varepsilon \|^2, \\
\mathbf{x}_t &= \sqrt{\bar{\alpha}_t} \mathbf{x} + \sqrt{1 - \bar{\alpha}_t} \varepsilon, \quad \varepsilon \sim \mathcal{N}(\mathbf{0}, \mathbf{I}).
\end{aligned}
\end{equation}

\noindent$\bullet$ \textbf{Idempotence Constraint}: To obtain consistency constraints on features, we adopt two types of idempotency losses, feature idempotency constraint and distribution idempotency constraint.

\noindent\textbf{1) Feature idempotency constraint} performs on features. We use the predicted noise to perform a step of de-noising to get the estimated generated data $\mathbf{x}_0$:
\begin{equation}
\mathbf{x}_0 = \frac{1}{\sqrt{\bar{\alpha}_t}} \left(\mathbf{x}_t - \sqrt{1 - \bar{\alpha}} g(\mathbf{x}_t, h(\mathbf{z}), t)\right).
\end{equation}
Therefore, the feature idempotency constraint based on this generated data $\mathbf{x}_0$ is formulated as:
\begin{equation}
\begin{aligned}
\mathcal{L}_{\text{ide\_feat}} &= -f(\mathbf{x})^T f(\mathbf{x}_0, \mathbf{z}_{t'}, t, t'),\\
\mathbf{z}_{t'} &= \sqrt{\bar{\alpha}_{t'}} \mathbf{z} + \sqrt{1 - \bar{\alpha}_{t'}} \varepsilon, \quad \varepsilon \sim \mathcal{N}(\mathbf{0}, \mathbf{I}).
\end{aligned}
\end{equation}
Since the generated data may be noisy, we input the features with noise and the number of time steps as auxiliary information.

\noindent\textbf{2) Distribution idempotency constraint} aims to align the feature distributions of the generated and original data. It is essential to maintain the manifold structure of the generated data consistent with the manifold of the original data. We capture the feature manifold structure through inter-feature similarity:
\begin{equation}
\begin{aligned}
\mathcal{P}(\mathbf{x}_0) &= f(\mathbf{x}_0)^T f(\mathbf{X}_0) = \left[f(\mathbf{x}_0)^T f(\mathbf{x}_0^1), \dots, f(\mathbf{x}_0)^T f(\mathbf{x}_0^m)\right], \\
\end{aligned}
\end{equation}
where $\mathbf{x}_0^i$ is $i$-th token data. We align it to the feature structure of the original ground truth data:
\begin{equation}
\mathcal{L}_{\text{ide\_dist}} = \mathcal{D}(\mathcal{P}(\mathbf{x}_0), \mathcal{P}(\mathbf{x})),
\end{equation}
where $\mathcal{D}(\cdot,\cdot)$ is the distance metric between two distributions. The feature idempotency constraint captures richer structural information and allows for the construction of tighter clusters. This is because the adjacency matrix not only connects different generated data of the same data but also connects different data with similar features. Based on this idempotent alignment, we enhance the generative power of the model for stronger perceptual performance, while the encoder learns stronger feature consistency. This results in reduced singular values $\sum_{i=d+1}^m \lambda_i^2$ of the adjacency matrix and better downstream task performance.

\section{Experiment Results}
To evaluate the effectiveness of our approach, we conducted experiments on two benchmark datasets: the NTU RGB+D dataset~\cite{shahroudy2016ntu,liu2019ntu} and the PKUMMD dataset~\cite{liu2020pku}.

\begin{table*}[tb]
  \scriptsize
  \centering
  \caption{Comparison of action recognition results with unsupervised learning approaches on NTU dataset.}
  \setlength{\tabcolsep}{3.0mm}{
  \begin{tabular}{lccccc}
    \toprule
   \multirow{2}{*}{Models}&\multirow{2}{*}{Architecture}&\multicolumn{2}{c}{NTU 60}&\multicolumn{2}{c}{NTU 120}\\
   \cmidrule{3-4} \cmidrule{5-6}
   & & xview & xsub & xset & xsub\\
    \midrule
    \rowcolor{gray!10} \multicolumn{6}{l}{\textit{Contrastive Learning:}}\\
    3s-AimCLR~\cite{guo2021contrastive} & GCN & 83.4 & 77.8 & 66.7 & 67.9\\
    3s-CPM~\cite{zhang2022contrastive} & GCN & 84.9 & 78.7 & 69.6 & 68.7\\
    3s-CMD~\cite{mao2022cmd} & GRU & 90.9 & 84.1 & 76.1 & 74.7 \\
    GL-Transformer~\cite{kim2022global} & Transformer & 83.8 & 76.3 & 68.7 & 66.0\\
    3s-ActCLR~\cite{lin2023actionlet} & GCN & 88.8 & 84.3 &  75.7 & 74.3\\
    \midrule
    \rowcolor{gray!10} \multicolumn{6}{l}{\textit{Generative Learning:}}\\
    3s-Colorization~\cite{yang2023self} & DGCNN & 87.2 & 79.1 & 70.8 & 69.2\\
    SkeletonMAE~\cite{yan2023skeletonmae} & GCN & 77.7 & 74.8 & 73.5 & 72.5\\
    MAMP~\cite{mao2023masked} & Transformer & 89.1 & 84.9 & 79.1 & 78.6\\
    \midrule
    \rowcolor{gray!10} \multicolumn{6}{l}{\textit{Contrative Learning} \& \textit{Generative Learning:}}\\
    CRRL~\cite{wang2022contrast} & GRU & 73.8 & 67.6 & 57.0 & 56.2\\
    PCM$^3$~\cite{zhang2023prompted} & GRU & 90.4 & 83.9 & 77.5 & 76.3\\
    \midrule
    \textbf{IGM (Ours)} & Transformer & \textbf{91.2} & \textbf{86.2}& \textbf{81.4} & \textbf{80.0}\\
    \bottomrule
\end{tabular}}
\label{tab:unsupervised_ntu}
\end{table*}

\begin{table}[t]
%
\scriptsize
\noindent
\begin{minipage}[t]{0.43\textwidth}
\centering
\captionof{table}{Comparison of action recognition results under KNN evaluation on NTU 60.}
\label{table:knn}
\setlength{\tabcolsep}{1.5mm}{
\begin{tabular}{lcc}
\toprule
Models &xview&xsub\\
\midrule
\rowcolor{gray!10} \multicolumn{3}{l}{\textit{Contrastive Learning:}}\\
AimCLR~\cite{guo2021contrastive} & 71.0 & 63.7 \\
SkeleMixCLR~\cite{chen2022contrastive} & 72.3 & 65.5 \\

\midrule
\rowcolor{gray!10} \multicolumn{3}{l}{\textit{Generative Learning:}}\\
LongT GAN~\cite{zheng2018unsupervised} & 48.1 & 39.1\\
MAMP~\cite{mao2023masked} & 70.0 & 62.0 \\
\midrule
\textbf{IGM \textit{w/o} $\mathcal{L}_{\text{ide}}$} & 67.2 & 64.7\\
\textbf{IGM \textit{w/} $\mathcal{L}_{\text{ide\_feat}}$} & 70.7 &68.4\\
\textbf{IGM \textit{w/} $\mathcal{L}_{\text{ide\_dist}}$} & 72.1 &69.0\\
\textbf{IGM (Ours)} & \textbf{72.6} &\textbf{69.3}\\
\bottomrule
\end{tabular}}

\end{minipage}%
\hfill
\begin{minipage}[t]{0.53\textwidth}
\centering
\captionof{table}{Comparison of the transfer learning performance on PKUMMD II dataset with linear evaluation pretrained on NTU 60.}
\label{table:transfer}
\setlength{\tabcolsep}{1.5mm}{
  \begin{tabular}{lcc}
  \toprule
   Models& xview& xsub\\
    \midrule
    \rowcolor{gray!10} \multicolumn{3}{l}{\textit{Finetune:}}\\
    LongT GAN~\cite{zheng2018unsupervised} & - & 44.8 \\
    MS$^2$L~\cite{lin2020ms2l} & - & 45.8\\
    ISC~\cite{thoker2021skeleton} & - & 51.1\\
    Hi-TRS~\cite{chen2022hierarchically} & - & 55.0\\
    3s-CrosSCLR~\cite{li20213d} & - & 51.3\\
    3s-AimCLR~\cite{guo2021contrastive} & 42.4 & 51.6 \\
    \midrule
    \rowcolor{gray!10} \multicolumn{3}{l}{\textit{Linear:}}\\
    3s-ActCLR~\cite{lin2023actionlet} & 44.5 & 55.9\\
    MAMP~\cite{mao2023masked} & 42.0 & 53.0 \\
    \textbf{IGM (Ours)} & \textbf{45.3} &\textbf{59.8}\\
    \bottomrule
\end{tabular}}
\end{minipage}
\end{table}

\begin{table*}[tb]
\scriptsize
\begin{center}
\caption{Action recognition accuracy for corruptions of test-time adaptation with single domain shift on NTU-C 60 xsub dataset.}
\label{tab:stta_all}
\setlength{\tabcolsep}{1.5mm}{
\begin{tabular}{lcccc}
\toprule
\multirow{2}{*}{Method} & \multicolumn{3}{c}{Joint Noise ($p$, $\sigma^2$)} & \multicolumn{1}{c}{Part Occlusion}\\
\cmidrule{2-4}\cmidrule{5-5}
& (1.0, 0.1) & (1.0, 0.05) & (0.5, 0.1) & Right Arms\\
\midrule
AimCLR~\cite{guo2021contrastive}&6.3&16.6&22.0&28.1\\
ActCLR~\cite{lin2023actionlet} &12.7&33.5&28.6&30.3\\
MAMP~\cite{mao2023masked} &2.4&6.1&5.8&10.7\\
\textbf{IGM (Ours)}&\textbf{58.7}&\textbf{63.0}&\textbf{65.3}&\textbf{56.9}\\
\midrule
\end{tabular}}
\end{center}
\end{table*}

\subsection{Datasets and Settings}
\noindent\textbf{NTU RGB+D Dataset 60 (NTU 60)}~\cite{shahroudy2016ntu} comprises a comprehensive compilation of 56,578 videos, covering 60 unique action labels. Every video includes annotations detailing the positions of 25 joints for each body, illustrating interactions among pairs and individual activities.

\noindent\textbf{NTU RGB+D Dataset 120 (NTU 120)}~\cite{liu2019ntu} is one of the most comprehensive datasets for recognizing actions. Encompassing 114,480 videos, it spans 120 unique action categories. This dataset documents the performance of actions by 106 individuals across diverse environments, employing 32 distinct recording configurations.

\noindent\textbf{PKU Multi-Modality Dataset (PKUMMD)}~\cite{liu2020pku} encompasses 52 action classes and nearly 20,000 instances, with each sample comprising 25 joints, thoroughly tackling the multi-modal 3D comprehension of human actions. The dataset is partitioned into two segments, with Part II showcasing more demanding data owing to heightened view diversity, resulting in skeleton noise.

For enhancing network training, all skeleton sequences undergo temporal downsampling to 120 frames. The encoder $f(\cdot)$ and generator $g(\cdot)$ are built using the Transformer architecture~\cite{yan2018spatial}, employing hidden channels configured to a dimension of 256. To assess performance, we employ a fully connected layer $\phi(\cdot)$.

To refine our network, we employ the Adam optimizer~\cite{newey1988adaptive}. Training is executed on a single NVIDIA GeForce RTX 4090, employing a batch size of 128, and the network undergoes training for 400 epochs.

\subsection{Evaluation and Comparison}
For a comprehensive assessment, we conduct comparative analysis of our approach with other methodologies across diverse scenarios.


\noindent\textbf{Linear Evaluation.}
In the linear evaluation framework, we utilize an encoder $f(\cdot)$ to process the extracted features and a linear classifier $\phi(\cdot)$ for action classification. The evaluation metric employed is the accuracy of action recognition. Notably, the encoder $f(\cdot)$ remains unchanged throughout the linear evaluation protocol. Our model demonstrates superior performance on the datasets outlined in Table~\ref{tab:unsupervised_ntu} compared to other methodologies.




\noindent\textbf{KNN Evaluation.}
In the K-Nearest Neighbors (KNN) evaluation setup, where the fixed encoder $f_q(\cdot)$ extracts features without any trainable parameters, our model showcases superiority in action recognition accuracy on the presented datasets. Table~\ref{table:knn} highlights the effectiveness of our approach compared to other methods in this evaluation mechanism.

\noindent\textbf{Transfer Learning.} 
In the transfer learning scenario, we assess the generalization capability of our model by pretraining it on the source data using a self-supervised task. We then evaluate the model's performance on the target dataset using the linear evaluation mechanism, with the encoder $f(\cdot)$ maintaining fixed parameters without additional fine-tuning. Our approach demonstrates superior performance in the transfer learning setting, as illustrated in Table~\ref{table:transfer}.

\begin{table}[t]
  \scriptsize
  \centering
  \caption{Comparison of mask prediction results on NTU 60 xsub.}
  \label{table:generation}
  \setlength{\tabcolsep}{3.0mm}{
  \begin{tabular}{lcccc}
    \toprule
    Method & DDPM~\cite{ho2020denoising} & MDM~\cite{tevet2022human} & SkeletonMAE~\cite{yan2023skeletonmae} & \textbf{IGM (Ours)} \\
    \midrule
    MPJPE (mm)$\downarrow$ & 130.2 & 87.6& 329.7 & \textbf{79.2}\\
    FID$\downarrow$ & 1.78 & 1.26 & 2.69 & \textbf{1.18}\\
    \bottomrule
  \end{tabular}
  }
\end{table}

\noindent\textbf{Zero-Shot Domain Generalization.}
By applying 4 types of corruption to the validation sets of all datasets, we assess the generalization of our proposed method compared to baseline approaches. For joint noise, we add noise with a probability of $p$ with a variance of $\sigma^2$ to some joints. We leverage the generative capability of our model, enabling us to denoise noisy skeleton data at test time. Subsequently, we utilize the generated skeleton data for recognition, significantly enhancing the generalization ability of our model. In Table~\ref{tab:stta_all}, our proposed approach shows consistent and substantial performance improvements. 

\begin{figure*}[t]
\centering
\begin{minipage}[t]{0.45\textwidth}
\centering
\includegraphics[width=\linewidth,height=0.7\textwidth]{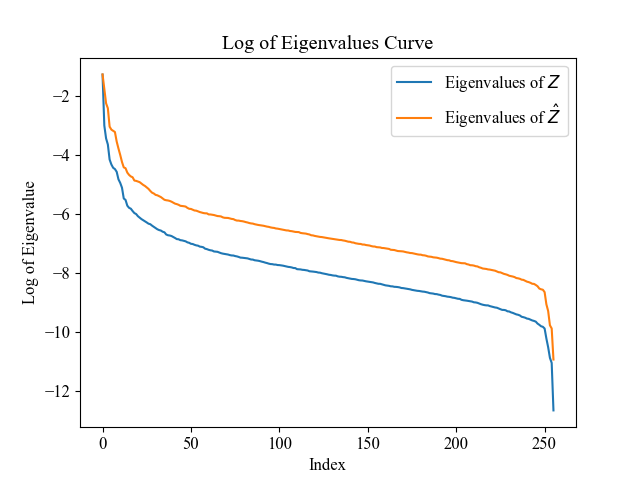}
\caption{
Curve of singular values with the singular value index.
}
\label{fig:eig}
\end{minipage}
\hspace{1mm}
\begin{minipage}[t]{0.45\textwidth}
\centering
\includegraphics[width=\linewidth,height=0.7\textwidth]{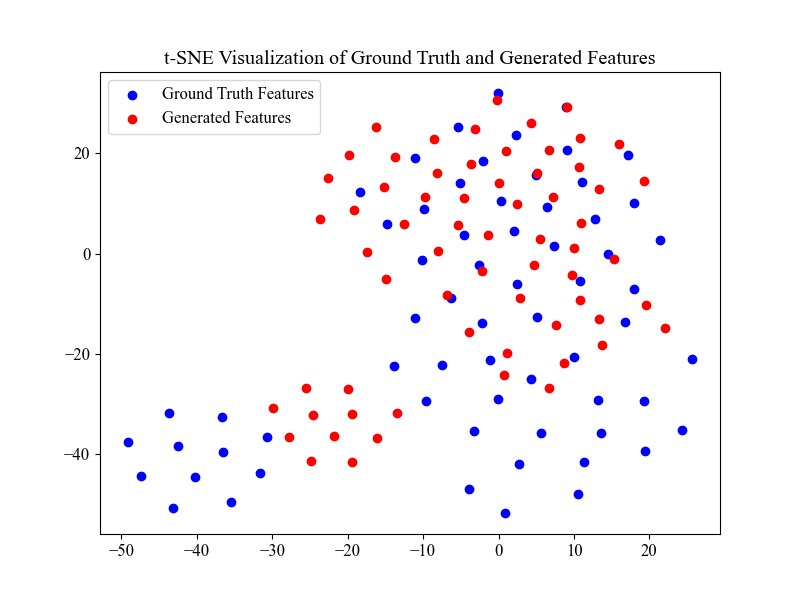}
\caption{
Visualisation of features in ground truth data and generated data.
}
\label{fig:feat}
\end{minipage}
\end{figure*}

\begin{figure*}[t]
\centering
\centering
\includegraphics[width=\linewidth]{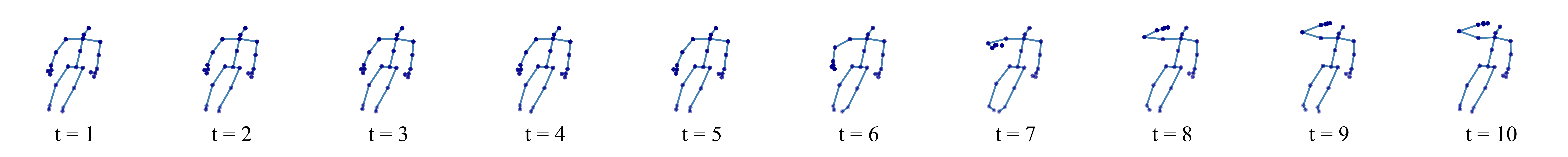}
\includegraphics[width=\linewidth]{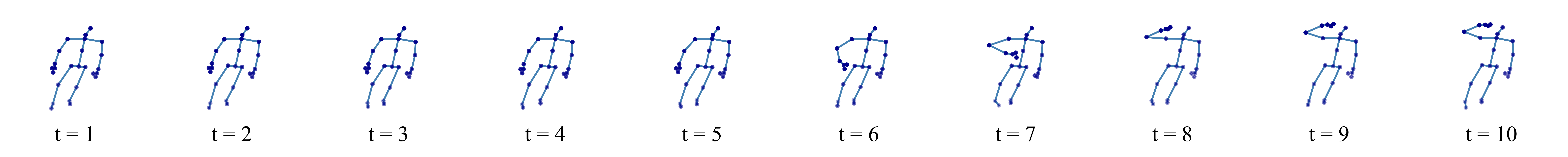}
\caption{
Visualisations of ground truth data and generated data. Above is the ground truth data, and below is the generated data. The conditions provided by the encoder are incorporated with data transformation, resulting in generated data that maintain similar motion information while exhibiting some diversity.
}
\label{fig:gen}
\end{figure*}

\noindent\textbf{Reconstruction Evaluation.}
In this section, we implement IGM for mask prediction tasks. We input the masked data into the encoder to extract features as conditions for generation, noting that the reconstruction task does not require adding data transformations to the conditions. Our method is compared with diffusion-based methods DDPM and MDM in Table~\ref{table:generation}. Figs~\ref{fig:feat} and~\ref{fig:gen} show visualizations and feature visualizations of both the generated data and the ground truth data. Despite sharing the same feature distribution, the generated samples exhibit some diversity due to the noise introduced in the conditions.

\begin{table}[t]
\scriptsize
\begin{center}
\caption{Analysis of module combinations on NTU 60 xsub dataset with the joint stream. ``FFM'' means Feature Fusion Module.}
\label{tab:md}
\setlength{\tabcolsep}{3.0mm}{
\begin{tabular}{c|c|c|c|c}
\toprule
\multicolumn{3}{c|}{Module}&\multirow{2}*{KNN}&\multirow{2}*{Linear}\\
\cmidrule{1-3}
FFM $h(\cdot)$ & $\mathcal{L}_{\text{ide\_feat}}$ & $\mathcal{L}_{\text{ide\_dist}}$ & ~ & ~\\
\midrule
 &  & & 64.7 & 83.3\\
$\checkmark$ &  & & 67.6 & 85.1\\
$\checkmark$ & $\checkmark$ & & 68.4 & 85.5\\
$\checkmark$ & & $\checkmark$ & 69.0 &  86.0 \\
$\checkmark$ &$\checkmark$& $\checkmark$ & \textbf{69.3} & \textbf{86.2}\\
\bottomrule
\end{tabular}}
\end{center}
\end{table}

\subsection{Ablation Study}
Here's the modified text for the ablation experiments:

\noindent\textbf{Analysis of Module Combination.} We investigate the performance of various combinations of modules and observe that each module contributes to a certain degree of improvement. Optimal performance is attained when all three modules are combined. As depicted in Table~\ref{tab:md}, each module enhances performance. 

\noindent\textbf{Analysis of Mitigating Dimensional Collapse.} The analysis points out that the feature space of the generated model is susceptible to dimensionality collapse, resulting in the extracted features losing the information needed for recognition. Fig.~\ref{fig:eig} shows the feature space of the encoder trained using the generative model and the feature space after Adapter. The token after removing similarity by Adapter network has higher feature values, \textit{i.e.}, the dimension collapse phenomenon is mitigated.

\section{Conclusions}
In this research, we propose the skeleton-based idempotent generative model (IGM) for unsupervised representation learning, presenting a novel framework that maximizes the potential of generative models for representation learning. By implementing idempotence at both the feature level and distribution, our model enriches features with semantic information about motion, making them more suitable for recognition tasks. Additionally, as the generative model primarily focuses on the principal component space, it is more susceptible to dimensional collapse. Conversely, recognition tasks rely more on the bottom subspace. To address this imbalance, we design an adapter that fuses encoder and generator features from different subspaces, thereby enhancing the effective feature dimension of the feature space.

\section*{Acknowledgements}
This work was supported in part by the National Natural Science Foundation of China under Grant No.62172020, and in part by the Key Laboratory of Science, Technology and Standard in Press Industry (Key Laboratory of Intelligent Press Media Technology).

\bibliographystyle{splncs04}
\bibliography{main}

\begin{thebibliography}{10}
\providecommand{\url}[1]{\texttt{#1}}
\providecommand{\urlprefix}{URL }
\providecommand{\doi}[1]{https://doi.org/#1}

\bibitem{bachman2019learning}
Bachman, P., Hjelm, R.D., Buchwalter, W.: Learning representations by maximizing mutual information across views. In: Proc.~Advances in Neural Information Processing Systems (2019)

\bibitem{chen2020simple}
Chen, T., Kornblith, S., Norouzi, M., Hinton, G.: A simple framework for contrastive learning of visual representations. In: Proc.~Int'l Conference for Machine Learning (2020)

\bibitem{chen2022hierarchically}
Chen, Y., Zhao, L., Yuan, J., Tian, Y., Xia, Z., Geng, S., Han, L., Metaxas, D.N.: Hierarchically self-supervised transformer for human skeleton representation learning. In: Proc.~European Conference on Computer Vision (2022)

\bibitem{chen2021channel}
Chen, Y., Zhang, Z., Yuan, C., Li, B., Deng, Y., Hu, W.: Channel-wise topology refinement graph convolution for skeleton-based action recognition. In: Proc.~Int'l Conference on Computer Vision (2021)

\bibitem{chen2022contrastive}
Chen, Z., Liu, H., Guo, T., Chen, Z., Song, P., Tang, H.: Contrastive learning from spatio-temporal mixed skeleton sequences for self-supervised skeleton-based action recognition. arXiv:2207.03065  (2022)

\bibitem{cheng2020skeleton}
Cheng, K., Zhang, Y., He, X., Chen, W., Cheng, J., Lu, H.: Skeleton-based action recognition with shift graph convolutional network. In: Proc.~IEEE Conference on Computer Vision and Pattern Recognition. pp. 183--192 (2020)

\bibitem{du2023role}
Du, T., Wang, Y., Wang, Y.: On the role of discrete tokenization in visual representation learning. In: Proc.~Int'l Conference on Learning Representations (2023)

\bibitem{erhan2010does}
Erhan, D., Bengio, Y., Courville, A., Manzagol, P.A., Vincent, P., Bengio, S.: Why does unsupervised pre-training help deep learning? Journal of Machine Learning Research  \textbf{11}(Feb),  625--660 (2010)

\bibitem{feichtenhofer2022masked}
Feichtenhofer, C., Li, Y., He, K., et~al.: Masked autoencoders as spatiotemporal learners. Proc.~Advances in Neural Information Processing Systems  \textbf{35},  35946--35958 (2022)

\bibitem{goutsu2015motion}
Goutsu, Y., Takano, W., Nakamura, Y.: Motion recognition employing multiple kernel learning of fisher vectors using local skeleton features. In: Proc.~Int'l Conference for Machine Learning Workshops (2015)

\bibitem{guo2021contrastive}
Guo, T., Liu, H., Chen, Z., Liu, M., Wang, T., Ding, R.: Contrastive learning from extremely augmented skeleton sequences for self-supervised action recognition. Proc.~AAAI Conference on Artificial Intelligence  (2022)

\bibitem{guo2023contranorm}
Guo, X., Wang, Y., Du, T., Wang, Y.: Contranorm: A contrastive learning perspective on oversmoothing and beyond. arXiv preprint arXiv:2303.06562  (2023)

\bibitem{haochen2106provable}
HaoChen, J., Wei, C., Gaidon, A., Ma, T.: Provable guarantees for self-supervised deep learning with spectral contrastive loss, 2021. arXiv preprint arXiv:2106.04156

\bibitem{he2022masked}
He, K., Chen, X., Xie, S., Li, Y., Doll{\'a}r, P., Girshick, R.: Masked autoencoders are scalable vision learners. In: Proc.~IEEE Conference on Computer Vision and Pattern Recognition (2022)

\bibitem{ho2020denoising}
Ho, J., Jain, A., Abbeel, P.: Denoising diffusion probabilistic models. Proc.~Advances in Neural Information Processing Systems  \textbf{33},  6840--6851 (2020)

\bibitem{isola2015learning}
Isola, P., Zoran, D., Krishnan, D., Adelson, E.H.: Learning visual groups from co-occurrences in space and time. arXiv:1511.06811  (2015)

\bibitem{jang2018grasp2vec}
Jang, E., Devin, C., Vanhoucke, V., Levine, S.: Grasp2{V}ec: Learning object representations from self-supervised grasping. arXiv:1811.06964  (2018)

\bibitem{kim2022global}
Kim, B., Chang, H.J., Kim, J., Choi, J.Y.: Global-local motion transformer for unsupervised skeleton-based action learning. Proc.~European Conference on Computer Vision  (2022)

\bibitem{li20213d}
Li, L., Wang, M., Ni, B., Wang, H., Yang, J., Zhang, W.: 3{D} human action representation learning via cross-view consistency pursuit. In: Proc.~IEEE Conference on Computer Vision and Pattern Recognition (2021)

\bibitem{lin2020ms2l}
Lin, L., Song, S., Yang, W., Liu, J.: {MS2L}: Multi-task self-supervised learning for skeleton based action recognition. In: Proc.~ACM Int’l Conference on Multimedia (2020)

\bibitem{lin2023actionlet}
Lin, L., Zhang, J., Liu, J.: Actionlet-dependent contrastive learning for unsupervised skeleton-based action recognition. In: CVPR (2023)

\bibitem{liu2020pku}
Liu, J., Song, S., Liu, C., Li, Y., Hu, Y.: A benchmark dataset and comparison study for multi-modal human action analytics. ACM Trans. on Multimedia Computing, Communications, and Applications  (2020)

\bibitem{liu2019ntu}
Liu, J., Shahroudy, A., Perez, M., Wang, G., Duan, L.Y., Kot, A.C.: {NTU RGB+D 120}: A large-scale benchmark for 3d human activity understanding. IEEE Trans. on Pattern Analysis and Machine Intelligence  (2019)

\bibitem{liu2022self}
Liu, X., Wang, Z., Li, Y.L., Wang, S.: Self-supervised learning via maximum entropy coding. Proc.~Advances in Neural Information Processing Systems  \textbf{35},  34091--34105 (2022)

\bibitem{liu2020disentangling}
Liu, Z., Zhang, H., Chen, Z., Wang, Z., Ouyang, W.: Disentangling and unifying graph convolutions for skeleton-based action recognition. In: Proc.~IEEE Conference on Computer Vision and Pattern Recognition (2020)

\bibitem{lorre2020temporal}
Lorre, G., Rabarisoa, J., Orcesi, A., Ainouz, S., Canu, S.: Temporal contrastive pretraining for video action recognition. In: Proceedings of the IEEE/CVF winter conference on applications of computer vision. pp. 662--670 (2020)

\bibitem{lv2006recognition}
Lv, F., Nevatia, R.: Recognition and segmentation of 3-d human action using hmm and multi-class adaboost. In: Proc.~European Conference on Computer Vision (2006)

\bibitem{mao2023masked}
Mao, Y., Deng, J., Zhou, W., Fang, Y., Ouyang, W., Li, H.: Masked motion predictors are strong 3d action representation learners. In: Proc.~Int'l Conference on Computer Vision. pp. 10181--10191 (2023)

\bibitem{mao2022cmd}
Mao, Y., Zhou, W., Lu, Z., Deng, J., Li, H.: {CMD}: Self-supervised 3d action representation learning with cross-modal mutual distillation. Proc.~European Conference on Computer Vision  (2022)

\bibitem{newey1988adaptive}
Newey, W.K.: Adaptive estimation of regression models via moment restrictions. Journal of Econometrics  (1988)

\bibitem{owens2018audio}
Owens, A., Efros, A.A.: Audio-visual scene analysis with self-supervised multisensory features. In: Proc.~European Conference on Computer Vision (2018)

\bibitem{peng2020learning}
Peng, W., Hong, X., Chen, H., Zhao, G.: Learning graph convolutional network for skeleton-based human action recognition by neural searching. In: Proc.~AAAI Conference on Artificial Intelligence (2020)

\bibitem{rai2021cocon}
Rai, N., Adeli, E., Lee, K.H., Gaidon, A., Niebles, J.C.: Cocon: Cooperative-contrastive learning. In: Proc.~IEEE Conference on Computer Vision and Pattern Recognition. pp. 3384--3393 (2021)

\bibitem{rao2021augmented}
Rao, H., Xu, S., Hu, X., Cheng, J., Hu, B.: Augmented skeleton based contrastive action learning with momentum {LSTM} for unsupervised action recognition. Information Sciences  (2021)

\bibitem{shahroudy2016ntu}
Shahroudy, A., Liu, J., Ng, T.T., Wang, G.: {NTU RGB+D}: A large scale dataset for 3d human activity analysis. In: Proc.~IEEE Conference on Computer Vision and Pattern Recognition (2016)

\bibitem{shi2019skeleton}
Shi, L., Zhang, Y., Cheng, J., Lu, H.: Skeleton-based action recognition with directed graph neural networks. In: Proc.~IEEE Conference on Computer Vision and Pattern Recognition. pp. 7912--7921 (2019)

\bibitem{shi2019two}
Shi, L., Zhang, Y., Cheng, J., Lu, H.: Two-stream adaptive graph convolutional networks for skeleton-based action recognition. In: Proc.~IEEE Conference on Computer Vision and Pattern Recognition (2019)

\bibitem{shotton2013real}
Shotton, J., Sharp, T., Kipman, A., Fitzgibbon, A., Finocchio, M., Blake, A., Cook, M., Moore, R.: Real-time human pose recognition in parts from single depth images. Communications of the ACM  (2013)

\bibitem{si2019attention}
Si, C., Chen, W., Wang, W., Wang, L., Tan, T.: An attention enhanced graph convolutional lstm network for skeleton-based action recognition. In: Proc.~IEEE Conference on Computer Vision and Pattern Recognition (2019)

\bibitem{si2018skeleton}
Si, C., Jing, Y., Wang, W., Wang, L., Tan, T.: Skeleton-based action recognition with spatial reasoning and temporal stack learning. In: Proc.~European Conference on Computer Vision. pp. 103--118 (2018)

\bibitem{song2022constructing}
Song, Y.F., Zhang, Z., Shan, C., Wang, L.: Constructing stronger and faster baselines for skeleton-based action recognition. IEEE transactions on pattern analysis and machine intelligence  \textbf{45}(2),  1474--1488 (2022)

\bibitem{song2020stronger}
Song, Y., Zhang, Z., Shan, C., Wang, L.: Stronger, faster and more explainable: A graph convolutional baseline for skeleton-based action recognition. In: Proc.~ACM Int’l Conference on Multimedia (2020)

\bibitem{su2020predict}
Su, K., Liu, X., Shlizerman, E.: Predict \& cluster: Unsupervised skeleton based action recognition. In: Proc.~IEEE Conference on Computer Vision and Pattern Recognition (2020)

\bibitem{tao2015moving}
Tao, L., Vidal, R.: Moving poselets: A discriminative and interpretable skeletal motion representation for action recognition. In: Proc.~Int'l Conference for Machine Learning Workshops (2015)

\bibitem{tevet2022human}
Tevet, G., Raab, S., Gordon, B., Shafir, Y., Cohen-Or, D., Bermano, A.H.: Human motion diffusion model. arXiv preprint arXiv:2209.14916  (2022)

\bibitem{thoker2021skeleton}
Thoker, F.M., Doughty, H., Snoek, C.G.: Skeleton-contrastive 3{D} action representation learning. In: Proc.~ACM Int’l Conference on Multimedia (2021)

\bibitem{tian2019contrastive}
Tian, Y., Krishnan, D., Isola, P.: Contrastive multiview coding. arXiv:1906.05849  (2019)

\bibitem{tian2020contrastive}
Tian, Y., Krishnan, D., Isola, P.: Contrastive multiview coding. In: Proc.~European Conference on Computer Vision. pp. 776--794. Springer (2020)

\bibitem{vemulapalli2014human}
Vemulapalli, R., Arrate, F., Chellappa, R.: Human action recognition by representing 3d skeletons as points in a lie group. In: Proc.~IEEE Conference on Computer Vision and Pattern Recognition (2014)

\bibitem{vemulapalli2016rolling}
Vemulapalli, R., Chellapa, R.: Rolling rotations for recognizing human actions from 3d skeletal data. In: Proc.~IEEE Conference on Computer Vision and Pattern Recognition (2016)

\bibitem{wang2012mining}
Wang, J., Liu, Z., Wu, Y., Yuan, J.: Mining actionlet ensemble for action recognition with depth cameras. In: Proc.~IEEE Conference on Computer Vision and Pattern Recognition (2012)

\bibitem{wang2016graph}
Wang, P., Yuan, C., Hu, W., Li, B., Zhang, Y.: Graph based skeleton motion representation and similarity measurement for action recognition. In: Proc.~European Conference on Computer Vision. pp. 370--385 (2016)

\bibitem{wang2022contrast}
Wang, P., Wen, J., Si, C., Qian, Y., Wang, L.: Contrast-reconstruction representation learning for self-supervised skeleton-based action recognition. IEEE Trans. on Image Processing  \textbf{31},  6224--6238 (2022)

\bibitem{wang2023message}
Wang, Y., Zhang, Q., Du, T., Yang, J., Lin, Z., Wang, Y.: A message passing perspective on learning dynamics of contrastive learning. arXiv preprint arXiv:2303.04435  (2023)

\bibitem{wei2022masked}
Wei, C., Fan, H., Xie, S., Wu, C.Y., Yuille, A., Feichtenhofer, C.: Masked feature prediction for self-supervised visual pre-training. In: Proc.~IEEE Conference on Computer Vision and Pattern Recognition. pp. 14668--14678 (2022)

\bibitem{wu2018unsupervised}
Wu, Z., Xiong, Y., Yu, S.X., Lin, D.: Unsupervised feature learning via non-parametric instance discrimination. In: Proc.~IEEE Conference on Computer Vision and Pattern Recognition (2018)

\bibitem{xu2024idempotence}
Xu, T., Zhu, Z., He, D., Li, Y., Guo, L., Wang, Y., Wang, Z., Qin, H., Wang, Y., Liu, J., et~al.: Idempotence and perceptual image compression. arXiv preprint arXiv:2401.08920  (2024)

\bibitem{yan2023skeletonmae}
Yan, H., Liu, Y., Wei, Y., Li, Z., Li, G., Lin, L.: Skeletonmae: graph-based masked autoencoder for skeleton sequence pre-training. In: Proc.~Int'l Conference on Computer Vision. pp. 5606--5618 (2023)

\bibitem{yan2018spatial}
Yan, S., Xiong, Y., Lin, D.: Spatial temporal graph convolutional networks for skeleton-based action recognition. In: Proc.~AAAI Conference on Artificial Intelligence (2018)

\bibitem{yang2021skeleton}
Yang, S., Liu, J., Lu, S., Er, M.H., Kot, A.C.: Skeleton cloud colorization for unsupervised 3{D} action representation learning. In: Proc.~Int'l Conference on Computer Vision (2021)

\bibitem{yang2023self}
Yang, S., Liu, J., Lu, S., Hwa, E.M., Hu, Y., Kot, A.C.: Self-supervised 3d action representation learning with skeleton cloud colorization. IEEE Trans. on Pattern Analysis and Machine Intelligence  (2023)

\bibitem{ye2019unsupervised}
Ye, M., Zhang, X., Yuen, P.C., Chang, S.F.: Unsupervised embedding learning via invariant and spreading instance feature. In: Proc.~IEEE Conference on Computer Vision and Pattern Recognition (2019)

\bibitem{yu2020learning}
Yu, Y., Chan, K.H.R., You, C., Song, C., Ma, Y.: Learning diverse and discriminative representations via the principle of maximal coding rate reduction. Proc.~Advances in Neural Information Processing Systems  \textbf{33},  9422--9434 (2020)

\bibitem{zhang2022contrastive}
Zhang, H., Hou, Y., Zhang, W., Li, W.: Contrastive positive mining for unsupervised 3d action representation learning. Proc.~European Conference on Computer Vision  (2022)

\bibitem{zhang2023prompted}
Zhang, J., Lin, L., Liu, J.: Prompted contrast with masked motion modeling: Towards versatile 3d action representation learning. In: Proc.~ACM Int’l Conference on Multimedia. pp. 7175--7183 (2023)

\bibitem{zhang2020semantics}
Zhang, P., Lan, C., Zeng, W., Xing, J., Xue, J., Zheng, N.: Semantics-guided neural networks for efficient skeleton-based human action recognition. In: Proc.~IEEE Conference on Computer Vision and Pattern Recognition (2020)

\bibitem{zhang2022mask}
Zhang, Q., Wang, Y., Wang, Y.: How mask matters: Towards theoretical understandings of masked autoencoders. Proc.~Advances in Neural Information Processing Systems  \textbf{35},  27127--27139 (2022)

\bibitem{zhang2020context}
Zhang, X., Xu, C., Tao, D.: Context aware graph convolution for skeleton-based action recognition. In: Proc.~IEEE Conference on Computer Vision and Pattern Recognition (2020)

\bibitem{zheng2018unsupervised}
Zheng, N., Wen, J., Liu, R., Long, L., Dai, J., Gong, Z.: Unsupervised representation learning with long-term dynamics for skeleton based action recognition. In: Proc.~AAAI Conference on Artificial Intelligence (2018)

\end{thebibliography}
\end{document}